\documentclass[final]{cvpr}

\usepackage{times}
\usepackage{epsfig}
\usepackage{graphicx}
\usepackage{amsmath}
\usepackage{amssymb}
\usepackage{booktabs}
\usepackage{multirow}
\usepackage[dvipsnames]{xcolor}
\usepackage[font=footnotesize,labelfont=bf]{caption}

\usepackage[pagebackref=true,breaklinks=true,letterpaper=true,colorlinks,bookmarks=false,citecolor=ForestGreen]{hyperref}

\def\cvprPaperID{XXXX} \def\confYear{CVPR 2021}

\renewcommand{\paragraph}[1]{\vspace{0.07in}\noindent {\bf #1}}

\begin{document}

\title{Self-supervised Pretraining of Visual Features in the Wild}

\author{
Priya Goyal$^1$
~~
Mathilde Caron$^{1, 2}$
~~
Benjamin Lefaudeux$^1$
~~
Min Xu$^1$
~~
Pengchao Wang$^1$
~~
Vivek Pai$^1$
\\
Mannat Singh$^1$
~~
Vitaliy Liptchinsky$^1$
~~
Ishan Misra$^1$
~~
Armand Joulin$^1$
~~
Piotr Bojanowski$^1$
\\
\\
$^1$ Facebook AI Research ~~~ $^2$ Inria\thanks{Univ. Grenoble Alpes, Inria, CNRS, Grenoble INP, LJK, 38000 Grenoble, France}
\\
\small Code: \url{https://github.com/facebookresearch/vissl}
}

\maketitle

\begin{abstract}
Recently, self-supervised learning methods like MoCo~\cite{he2020momentum}, SimCLR~\cite{chen2020simple}, BYOL~\cite{grill2020bootstrap} and SwAV~\cite{caron2020unsupervised} have reduced the gap with supervised methods.
These results have been achieved in a control environment, that is the highly curated ImageNet dataset. 
However, the premise of self-supervised learning is that it can learn from any random image and from any unbounded dataset. 
In this work, we explore if self-supervision lives to its expectation by training large models on random, uncurated images with no supervision.
Our final SElf-supERvised (SEER) model, a RegNetY with 1.3B parameters trained on 1B random images with 512 GPUs achieves 84.2\% top-1 accuracy, surpassing the best self-supervised pretrained model by 1\% and confirming that self-supervised learning works in a real world setting.
Interestingly, we also observe that self-supervised models are good few-shot learners achieving 77.9\% top-1 with access to only 10\% of ImageNet. 
\end{abstract}

\section{Introduction}

A recent trend shows that well-tailored model pretraining approaches (weakly-supervised, semi-supervised, self-supervised) can drastically improve the performance on downstream tasks for most deep learning applications.
It has been observed for Natural Language Processing~\cite{devlin2018bert,radford2019language}, Speech Recognition~\cite{riviere2020unsupervised,schneider2019wav2vec} and Computer Vision~\cite{mahajan2018exploring}.
There are two key ingredients that have contributed towards this success.
The \emph{first} is pretraining on massive datasets:
the GPT-3~\cite{brown2020language} language model is pretrained on $300$B words, while the speech model Wav2vec2.0~\cite{baevski2020wav2vec} is learned on $53$K hours of audio~\cite{kahn2020libri}.
The \emph{second} ingredient is the use of models with massive capacity, even  reaching hundreds of billions of parameters for the largest NLP models~\cite{brown2020language,raffel2019exploring}.

\begin{figure}[t]
  \includegraphics[width=0.99\linewidth]{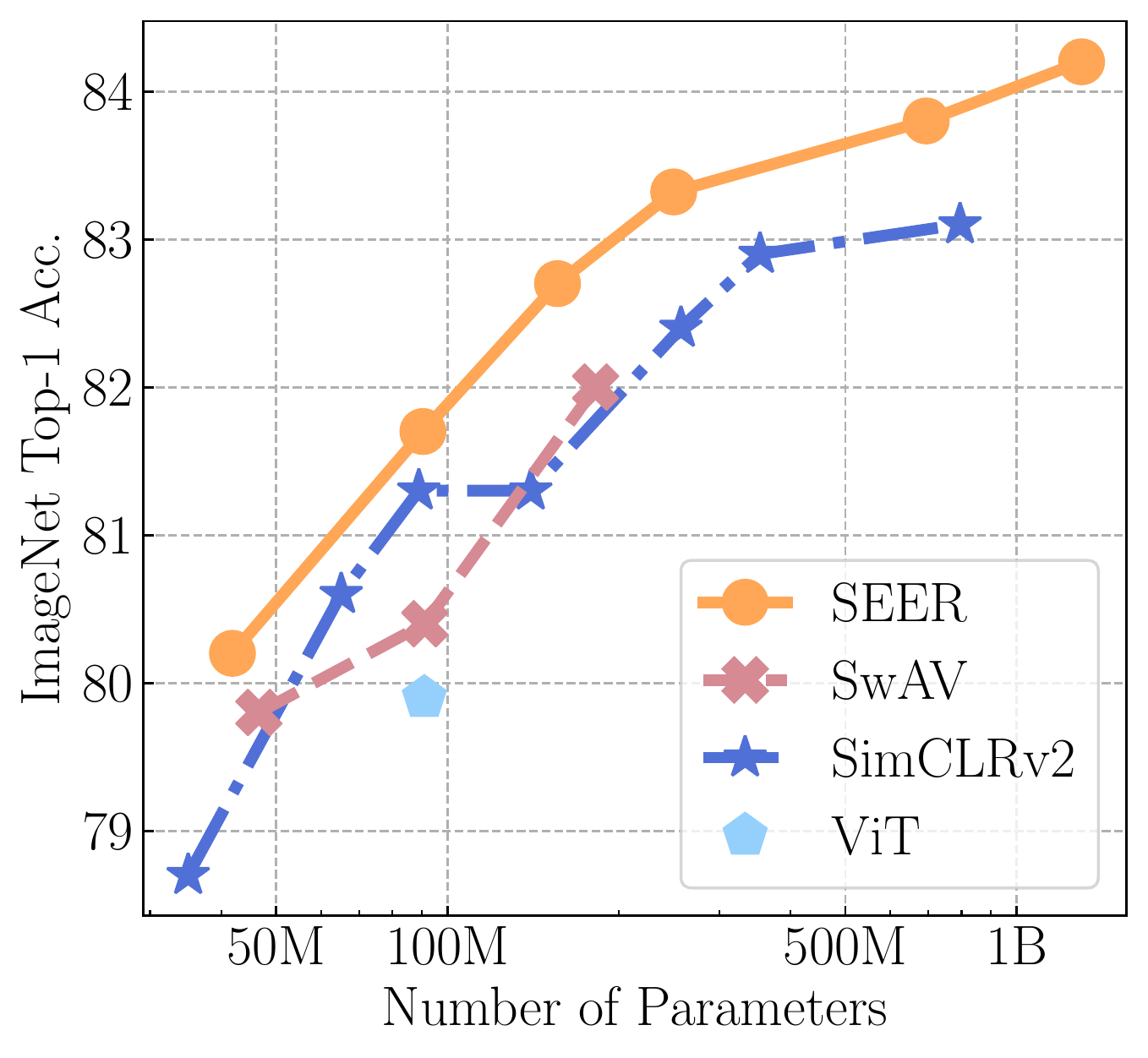} 
  \vspace{-0.1in}
  \caption{
    \textbf{Performance of large pretrained models on ImageNet.}
    We pretrain our SEER models n an uncurated and random images.
    They are RegNet architectures~\cite{radosavovic2020designing} trained with the SwAV self-supervised method~\cite{caron2020unsupervised}
    We compare with the original models trained in Caron et al.~\cite{caron2020unsupervised}
    as well as the pretraining on curated data from SimCLRv2~\cite{chen2020big} and ViT~\cite{dosovitskiy2020image}.
    The network architectures are different. 
    We report the top-1 accuracy after finetuning on ImageNet.
  }
  \label{fig:pullfig}
\end{figure}

While the benefit of pretraining has been demonstrated in computer vision, it has been in the limited scope of curated datasets originally collected for supervised or weakly supervised learning. These datasets represent only a limited fraction of the general distribution of internet scale images~\cite{henaff2019data,kolesnikov2019big,mahajan2018exploring}. 
Prior attempts to train self-supervised models on uncurated data~\cite{caron2019unsupervised,doersch2015unsupervised,goyal2019scaling,joulin2016learning} have only used a few millions of images for which using small model architectures is sufficient. This still leaves an open question - can we achieve good performance by pretraining on an \emph{extremely large} collection of random, uncurated and unlabeled images? Answering this question has important implications. 
Practically, it may lead to strategies for pretraining that use unlabeled data to achieve state-of-the-art performance on transfer learning, and to create systems that continually learn in a self-supervised manner from an unending data stream.

In this work, we address this question by pretraining high capacity models on billions of random internet images, \ie, completely unconstrained images from the internet. We do not rely on image meta-data or any form of weak/manual annotations to filter data or train the model.
Training powerful image representations on unlabeled data has recently been made possible with advances in self-supervised learning~\cite{caron2019unsupervised,chen2020simple,grill2020bootstrap,he2020momentum}.
The most recent self-supervised models pretrained on ImageNet~\cite{russakovsky2015imagenet} even surpass supervised pretrained models on multiple downstream tasks~\cite{he2020momentum}. 
Recent developments~\cite{caron2020unsupervised} have also shown their potential when trained on random internet images.
Further, these methods are amenable to online learning, making them perfect candidates for training large models on unlimited data.

For our analysis, we focus on the RegNet family of architectures~\cite{radosavovic2020designing} and in particular the architecture with $700$M parameters.
The RegNet architectures are particularly well suited for this task for two reasons. 
First, they offer an excellent trade-off of efficiency and performance. Second, they are very flexible for scaling the number of parameters.
We train these models online on a dataset of $2$B random internet images using the SwAV self-supervised approach~\cite{caron2020unsupervised}.
We use SwAV for this study for its fast convergence to good performance in the online setting with large batch size.
To make this study tractable at this scale, we leverage several existing tools to reduce the memory usage of our models, including mixed precision and gradient checkpointing.

The main finding of our study is that our SElf-supERvised (``SEER'') pretrained models are not only good for initializing training on curated dataset like ImageNet, they are also excellent few shot learners, achieving 75.1\% with only 10\% of ImageNet.
Our model also achieves better performance than supervised model trained on ImageNet on several downstream tasks, confirming the benefits of self-supervised pretraining, even when performed on uncurated data.

\section{Related Work}

Our work explores the limits of training large architectures on large uncurated datasets with self-supervised learning.
We build on prior work from different areas: self-supervised learning, training at scale and large convolutional network architectures.

\paragraph{Unsupervised pretraining of visual features.}
Self-supervised learning has a long history in computer vision with methods based on autoencoders~\cite{ranzato2007unsupervised,vincent2008extracting}, clustering~\cite{asano2019self,caron2018deep,coates2011analysis} or instance-level discrimination~\cite{bojanowski2017unsupervised, dosovitskiy2016discriminative,hadsell2006dimensionality,wu2018unsupervised}.
Recently, methods based on contrastive learning~\cite{hadsell2006dimensionality,oord2018representation} have shown that unsupervised pretraining produces features that surpass the supervised feature representations on many downstream tasks~\cite{caron2020unsupervised,chen2020simple,grill2020bootstrap,he2020momentum,misra2020self}.
These methods discriminate either between each instance feature~\cite{chen2020simple, he2020momentum, misra2020self} or between their cluster assignments~\cite{baevski2020wav2vec,caron2020unsupervised,li2020prototypical}.
Most works on unsupervised pretraining focus on supervised datasets like ImageNet~\cite{russakovsky2015imagenet} or curated datasets collected by filtering images related to pre-defined labels~\cite{sun2017revisiting}.
The key takeaway from these works is that supervised labels are not required as long as you trained on the filtered data.
Some works have explored unsupervised training in the wild for images~\cite{caron2019unsupervised,doersch2015unsupervised,goyal2019scaling} and videos~\cite{miech2020end}.
These studies were conducted at a small scale, and there are now evidences that self-supervised pretraining benefits greatly from large archtiectures~\cite{caron2020unsupervised,chen2020big, henaff2019data}.
Our work builds upon these findings to explore if we can learn good visual representations by training large architectures on large collection of random, uncurated and unlabeled images.

\paragraph{Learning of visual features at scale.}
Benefiting from the advances in distributed training~\cite{goyal2017accurate}, several works have shown the advantages of pretraining on large curated image datasets with weak-supervised learning~\cite{joulin2016learning, mahajan2018exploring}, semi-supervised learning~\cite{yan2020clusterfit} or supervised training on hundreds of millions of filtered images~\cite{kolesnikov2019big,sun2017revisiting}.
Of particular interest, Mahajan et al.~\cite{mahajan2018exploring} show that the pretraining on billions of images significantly improves the performance of large architectures compared to training them from scratch.
Most works on training at large data scale rely on a data filtering step to only keep the images associated with targeted concepts.
This filtering either uses hastags that are synsets of ImageNet classes~\cite{mahajan2018exploring, yan2020clusterfit}, or the predictions from a pretrained object classifier~\cite{sun2017revisiting}.
As opposed to this line of work, we are interested in learning features that cover any available image, and hence, we do not curate our training dataset to match a pre-defined set of concepts.

\paragraph{Scaling architectures for image recognition.}
Many works have shown the benefits of training large architectures on the quality of the resulting visual features~\cite{radosavovic2020designing,tan2019efficientnet,xie2017aggregated}. 
Training large architectures is especially important when pretraining on a large dataset, where a model with limited capacity will underfit~\cite{mahajan2018exploring}.  
This becomes further more important when the pretraining is performed with contrastive learning, where the network has to learn to discriminate between each instance of the dataset~\cite{caron2019unsupervised,chen2020simple,chen2020big,grill2020bootstrap} in order to learn good visual representations.
For instance, Kolesnikov et al.~\cite{kolesnikov2019revisiting} have demonstrated the importance of training wider networks for the quality of visual features learned with self-supervision. 
More recently, Chen et al.~\cite{chen2020big} have achieved impressive performance with deeper and wider configurations of ResNet~\cite{he2016deep}.
However, scaling architectures for image recognition goes beyond simply changing the width and the depth of a model, and a large amount of literature is dedicated to building scale efficient models with large capacity~\cite{tan2019efficientnet,xie2017aggregated,touvron2020fixing}.
Of particular interest, the RegNets~\cite{radosavovic2020designing} achieve competitive performance on standard image benchmarks while offering an efficient runtime and memory usage making them a candidate for training at scale. 
In our work, we show the benefits of this model family for large scale self-supervised pretraining.

\section{Method}

In this section, we provide a brief overview of the components used in this work to pretrain visual features in the wild.
We describe the self-supervised method, SwAV~\cite{caron2020unsupervised}, and the family of convnet architectures, RegNet~\cite{radosavovic2020designing}. 
We then discuss several technical strategies required to train large models on billions of images with self-supervision.

\subsection{Self-Supervised Pretraining}

We pretrain our model with an online self-supervised approach called SwAV that we briefly summarize in this section.
We refer to Caron et al.~\cite{caron2020unsupervised} for more details.

SwAV is an online clustering method to train convnets without annotations. It works by training an embedding that yields consistent cluster assignments between multiple views of the same image.
By mining clusters invariant to data augmentations, the system learns semantic representations.
In practice, SwAV works by comparing the features of different views of the same image using their intermediate cluster assignments.
If these features capture the same information, it should be possible to predict the assignment of one from the feature of another view.
More precisely, we consider a set of $K$ clusters, each associated with a learnable $d$-dimensional prototype vector $\mathbf{v}_k$.
Given a batch of $B$ images, each image $i$ is transformed into two views: $\mathbf{x}_{i1}$ and $\mathbf{x}_{i2}$.
All views are then featurized with a convnet, resulting in two sets of features $(\mathbf{f}_{11},\dots, \mathbf{f}_{B1})$ and $(\mathbf{f}_{12},\dots,\mathbf{f}_{B2})$.
Each set of features is assigned independently to the cluster prototypes using an Optimal Transport solver.
This solver enforces that the features are uniformly split across clusters, avoiding trivial solutions where all representations are mapped to an unique prototype.
The resulting assignments are then swapped between the two sets: the cluster assignment $\mathbf{y}_{i1}$ of the view $\mathbf{x}_{i1}$ has to be predicted from the feature representation $\mathbf{f}_{i2}$ of the view $\mathbf{x}_{i2}$, and vice-versa.
Formally, the convnet and prototypes weights are trained to minimize the following loss for all examples $i$:
\begin{eqnarray*}
L(\mathbf{f}_{i1}, \mathbf{f}_{i2}) & = &\ell(\mathbf{f}_{i1}, \mathbf{y}_{i2}) + \ell(\mathbf{f}_{i2}, \mathbf{y}_{i1}).
\end{eqnarray*}

The cluster prediction loss~$\ell({\mathbf f},{\mathbf y})$ is the cross entropy between the cluster assignment and a softmax of the dot products of $\mathbf{f}$ and all prototypes $\mathbf{v}_k$:
\begin{equation*}
  \ell(\mathbf{f}, \mathbf{y}) = - \sum_{k} \mathbf{y}^{(k)} \log \mathbf{p}^{(k)}
\end{equation*}
where:
\begin{equation*}
\mathbf{p}^{(k)} = \frac{ \exp \left ( \frac{1}{\tau} \mathbf{f}_t^\top \mathbf{v}_k \right ) }{\sum_{k'} \exp \left ( \frac{1}{\tau} \mathbf{f}_t^\top \mathbf{v}_{k'} \right ) }.
\end{equation*}

\subsection{Scale efficient model family: RegNetY}

Scaling data and model capacity jointly requires using architectures that are efficient in terms of both 
memory and runtime.
RegNets are a family of models designed for this purpose and we briefly describe them in this section. 
We refer to Radosavovic et al.~\cite{radosavovic2020designing} for more details.

RegNets are a family of architectures defined by a design space of convnets consisting of 4 stages, with each stage containing a series of identical blocks, while keeping the structure of their blocks fixed -- namely the residual bottleneck block of He et al.~\cite{he2016deep}. In this work, we focus on the RegNetY architectures, that add a Squeeze-and-excitation op~\cite{hu2018squeeze} to the standard RegNets to further improve their performance. The RegNetY model family is parameterized by 5 parameters, allowing the search of a good instance with a certain number of FLOPs with reasonable resources. The models we used were all searched on ImageNet using the same procedure as Radosavovic et al.~\cite{radosavovic2020designing}. We believe our results can further be improved by searching for RegNetYs directly on our self-supervised pre-training task.

\paragraph{The \texttt{RegNetY-256GF} architecture.}
Our model of focus is the \texttt{RegNetY-256GF} architecture. 
Its parametrization is given by the scaling rules of RegNets~\cite{radosavovic2020designing}: 

\begin{align*}
\small
    w_0 = 640, w_a = 230.83, w_m = 2.53, \text{group width} = 373
\end{align*}
It has 4 stages with stage depths ($2$, $7$, $17$, $1$) and stage widths ($528$, $1056$, $2904$, $7392$), leading to a total of $695.5$M parameters.
It takes $6125$ms for a single training iteration over $8,704$ images on $512$ V100 32GB NVIDIA GPUs.
Training this model on $1$ billion images requires $114,890$ training iterations for a batch size of $8,704$ images, summing to $8$ days of training over $512$ GPUs.

\subsection{Optimization and Training at Scale}

In this work, we propose several adjustments to the training of self-supervised methods to adapt it to a large scale.

\paragraph{Learning rate schedule.}
We explore two learning rate schedules: the cosine wave~\cite{loshchilov2016sgdr} and a simpler fixed learning rate schedule.
The cosine wave adapts to the number of updates and we focus on this scheduling for fair comparison between different models.
However, it is not adapted to online large scale training because it uses the total of updates for scheduling and it also weighs images differently depending on when they are seen during training.
For this reason, we also explore a fixed learning rate schedule.
In this scheduling, we keep the learning rate fixed until the loss is non-decreasing, then we divide the learning rate by $2$.
Our observation is that this schedule works as well in practice and allows for more flexible training.
However, we train our largest model, the \texttt{RegNetY-256GF}  with cosine learning rate schedule since we use only 1B images.

\paragraph{Reducing memory consumption per GPU.}
We reduce the amount of GPU memory required during training with gradient checkpointing~\cite{chen2016grad} and mixed precision. We use \texttt{O1} optimization level from NVIDIA Apex library\footnote{\url{https://github.com/NVIDIA/apex}} to perform operations like GEMMs and convolutions in $16$-bits floating-point precision.
We use PyTorch's gradient checkpointing implementation which trades compute for memory. 
It discards intermediate activations during the forward pass, and recomputes them during the backward pass. 
In our experiments, using gradient checkpointing, we observe negligible compute overhead in memory-bound settings.

\paragraph{Optimizing Training speed.}
Enabling mixed-precision for memory optimization has additional benefits, as modern accelerators take full advantage of the FP$16$ reduced size by increasing throughput when compared to FP$32$. This improves memory-bandwidth bottleneck and speeds up training.
We also use the optimized SyncBatchNorm implementation with kernels through CUDA/C++ extensions from NVIDIA Apex library. For synchronizing BatchNorm layer across GPUs, we create process groups instead of performing global sync which is slow. Finally, our dataloader pre-fetches more training batches leading to higher data throughput than the default PyTorch dataloader.

\paragraph{Large scale Pretraining data.}
For our billion scale pretraining, we consider a dataloader that directly samples random, public, and non-EU images from Instagram.
As we train online and in the wild, we do not apply any curation or pre-processing on the images, such as hashtag filtering or de-duplication.
This dataset is not static and gets refreshed every $90$ days, however, we can confirm that the refreshment doesn't degrade the model performance.

\paragraph{Implementation details.}

We pretrain a  \texttt{RegNetY-256GF} with SwAV, using $6$ crops per image of resolutions $2\times224 + 4\times96$. 
We follow the same data augmentation as in Caron et al.~\cite{caron2020unsupervised}.
During pretraining, we use a 3-layer multi-layer perceptron (MLP) projection head of dimensions $10444\times8192$, $8192\times8192$ and $8192\times256$. We do not use BatchNorm layers in the head.
We use $16$K prototypes, temperature $\tau$ set to $0.1$, the Sinkhorn regularization parameter $\epsilon$ to $0.05$ and perform $10$ iterations of Sinkhorn algorithm.
We synchronize BatchNorm stats across gpus and create process groups of size 64 for synchronization.
We use a weight decay of $10^{-5}$, LARS optimizer~\cite{you2017large} and \texttt{O1} mixed-precision optimization from Apex library. We also apply activation checkpointing~\cite{chen2016grad}.
We train our model with stochastic gradient descent using a large batch size of $8192$ different images distributed over $512$ NVIDIA V100 32GB GPUs, resulting in $16$ different images per GPU. 
The learning rate is linearly ramped up from $0.15$ to $9.6$ for the first $8$K training updates. 
After warmup, we follow a cosine learning rate schedule and decay the learning rate to final value $0.0096$.  
Overall, we train on $1$B images for a total of $122$K iterations.

\section{Main Results}

We study the quality of the features generated by our self-supervised pretraining on a variety of downstream tasks and benchmarks.
We also consider a low-shot setting with limited access to images and their labels for the downstream task, as well as, standard evaluation using the entire data available for the downstream task.
We also compare with prior work trained on large curated datasets.

\subsection{Finetuning Large Pretrained Models}

In this section, we measure the quality of models pretrained in the wild by transferring them to the ImageNet object classification benchmark.

\begin{table*}[t]
\centering
\begin{tabular}{ l l l l c c}
\toprule
 Method & Data & \#images& Arch. & \#param. & Top-1\\
\midrule
DeeperCluster~\cite{caron2019unsupervised} & YFCC100M & 96M & VGG16 & 138M & 74.9\\ 
ViT~\cite{dosovitskiy2020image} & JFT & 300M & ViT-B/16 & \phantom{0}91M & 79.9\\
SwAV~\cite{caron2020unsupervised} & IG & 1B & RX101-32x16d & 182M & 82.0 \\
SimCLRv2~\cite{chen2020big} & ImageNet & 1.2M & RN152w3+SK & 795M & 83.1\\
\midrule
SEER & IG & 1B & RG128 & 693M & 83.8\\
SEER & IG & 1B & RG256 & 1.3B & \textbf{84.2}\\
\bottomrule
\end{tabular}
\vspace{.2em}
\caption{
\textbf{Finetuning of models pretrained with self-supervision.}
We compare with existing features pretrained with no supervision.
After pretraining, the models are finetuned on ImageNet and we report top-1 accuracy.
We give the details of the architectures and datasets used for pretraining.
Numbers are taken from the respective papers.
DeepCluster and SwAV are pretrained on uncurated dataset, while SimCLRv2 is pretrained on ImageNet only, and ViT is pretrained on a curated dataset of 300M images.
}
\label{tab:finetune}
\end{table*}

\paragraph{Experimental setting.}
We pretrain $6$ RegNet architectures of different capacities, namely \texttt{RegNetY-\{8,16,32,64,128,256\}GF}, on $1$B random, public and non-EU Instagram images with SwAV.
We finetune these models on the task of image classification on ImageNet, using the standard $1.28$M training images with labels and evaluate on $50k$ images in the standard validation set.
We apply the same data augmentation as in SwAV~\cite{caron2020unsupervised}.
We finetune for $35$ epochs with SGD, batch size of $256$, learning rate of $0.0125$ reduced by factor of $10$ after $30$ epochs, weight decay of $10^{-4}$ and momentum of $0.9$.
We report top-1 accuracy on validation set using the $224\!\times\!224$ center crop.

\paragraph{Comparision with other self-supervised pretraining.}
In Table~\ref{tab:finetune}, we compare our largest pretrained model, a \texttt{RegNetY-256GF}, with existing self-supervised pretrained models. 
We achieve \textit{\textbf{84.2\% top-1 accuracy on ImageNet}}, surpassing by +1\%, the best existing pretrained model from SimCLRv2~\cite{chen2020big}.
In the Figure~\ref{fig:pullfig}, we show the same comparison with different model capacities.
The conclusion remains unchanged regardless of the model capacity, showing that combining RegNet with SwAV is a good candidate for pretraining.

\begin{figure}[t]
  \centering
  \includegraphics[width=.92\linewidth]{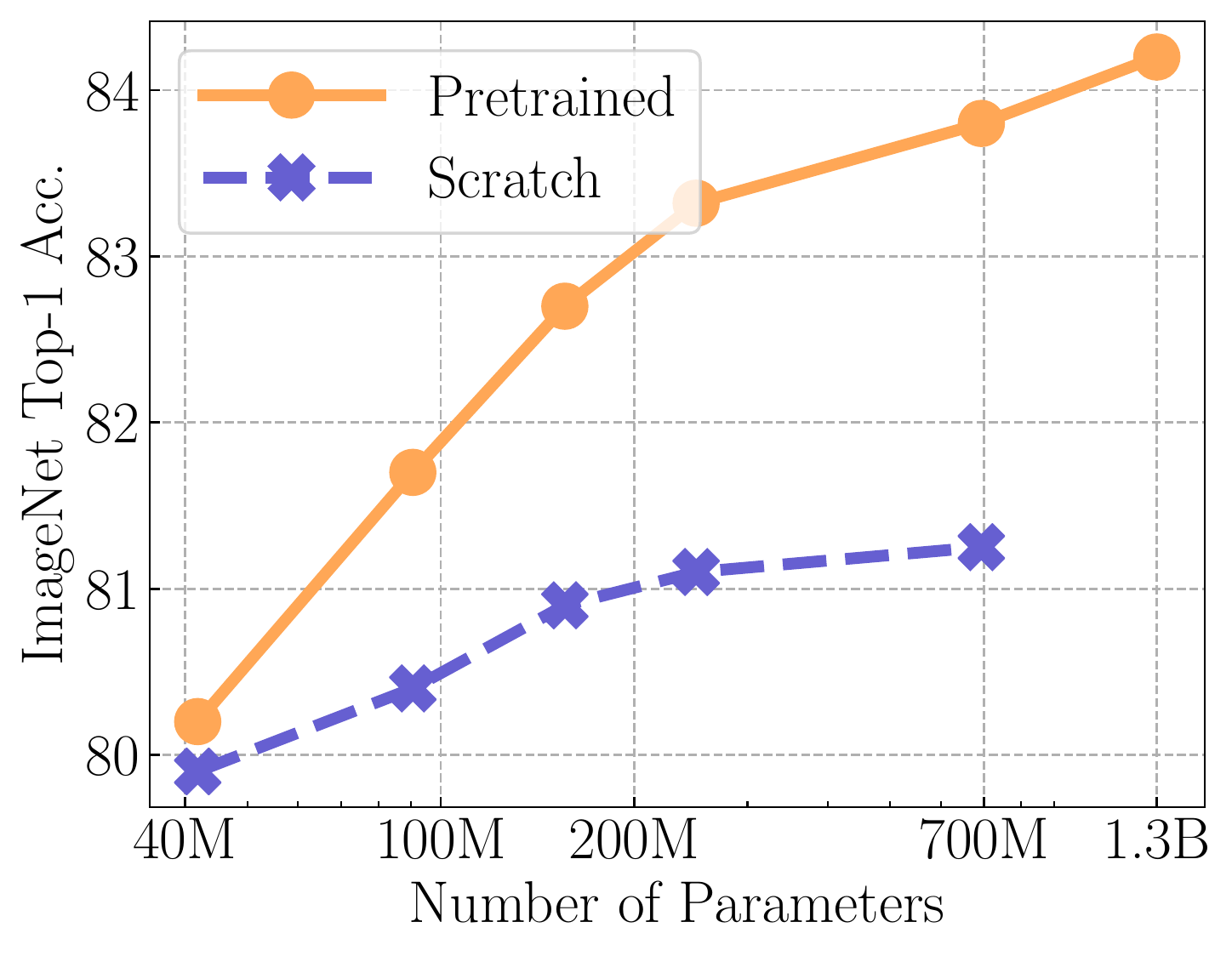} 
  \caption{
\textbf{Finetuning pretrained RegNets on ImageNet versus Scratch.} 
We show the impact of finetuning pretrained \texttt{RegNetY-\{8,16,32,64,128\}GF} compared to training them from scratch on ImageNet.
The pretrained RegNet models are trained with our self-supervised approach on $1$B random IG images.
We report top-1 accuracy on the validation set.
}
\label{fig:finetune}
\end{figure}

\paragraph{Impact of the model capacity.}
In Figure~\ref{fig:finetune}, we show the impact of model capacity on the performance of pretraining compared to training from scratch.
While model capacity benefits both initializations, it has a more significant impact on pretrained models when scaled to hundreds of millions of parameters.
A reason is that training these architecture from scratch could overfit on ImageNet which is a relatively small dataset.
We confirm that the log-scale performance gain from increasing model capacity also appears in the case where the pretraining data is uncurated.

\begin{table}[t]
  \centering
  \begin{tabular}{@{}ll c cc@{}}
    \toprule
Method  & Arch. & Param. & 1\% & 10\%  \\
    \midrule
Scratch      & RG128 & 848M & 12.8 & 54.5  \\
\midrule
\multicolumn{5}{@{}l}{\textit{Semi-supervised methods on full ImageNet}}\\
FixMatch~\cite{sohn2020fixmatch} & RN50 & \phantom{0}24M & - & 71.5 \\
CowMix~\cite{french2020milking}  & RN152 & 265M & - & 73.9 \\
\midrule
    \multicolumn{5}{@{}l}{\textit{Self-supervised pretraining on full ImageNet}}\\
SimCLR~\cite{chen2020simple}      & RN50   & \phantom{0}24M & 48.3 & 65.6 \\
SwAV~\cite{caron2020unsupervised} & RN50   & \phantom{0}24M & 53.9 & 70.2 \\
BYOL~\cite{grill2020bootstrap}	  & RN200  & 250M   & 71.2 & 77.7 \\ 
SimCLR v2~\cite{chen2020big}      & RN152w3+SK & 795M & 74.9  & 80.1 \\
\midrule
     \multicolumn{5}{@{}l}{\textit{Pretrained on random internet images}}\\
          SEER & RG128 & 693M & 57.5 & 76.7 \\
     SEER & RG256 & 1.3B & 60.5 & 77.9 \\
    \bottomrule
  \end{tabular}
\vspace{.3em}
  \caption{
    \textbf{Low-shot learning on ImageNet.}
We compare our approach with semi-supervised approaches and self-supervised pretraining on low-shot learning.
Our model is finetuned on either 1\% or 10\% of ImageNet, and \textit{does not access the rest of ImageNet images}.
As opposed to our method, the other methods use all the images from ImageNet during pretraining or finetuning.
  }
  \label{tab:semi_sup_inet}
\end{table}

\subsection{Low-shot learning}

In this section, we are interested in evaluating the performance of our pretrained model in the low-shot setting, i.e., with a fraction of data on the downstream task.

\paragraph{Experimental setting.}
We consider two datasets for low-shot learning, namely ImageNet~\cite{russakovsky2015imagenet} and Places205~\cite{zhou2014learning}.
We assume a limited access to the dataset during transfer learning, both in terms of labels \emph{and} images.
This setting differs from the standard setting used in self-supervised learning where the entire datasets is accessible and only the access to labels is limited~\cite{henaff2019data}.
For the rest, we follow their experimental setting for finetuning the features.

\begin{figure}[t]
  \centering
\includegraphics[width=.92\linewidth]{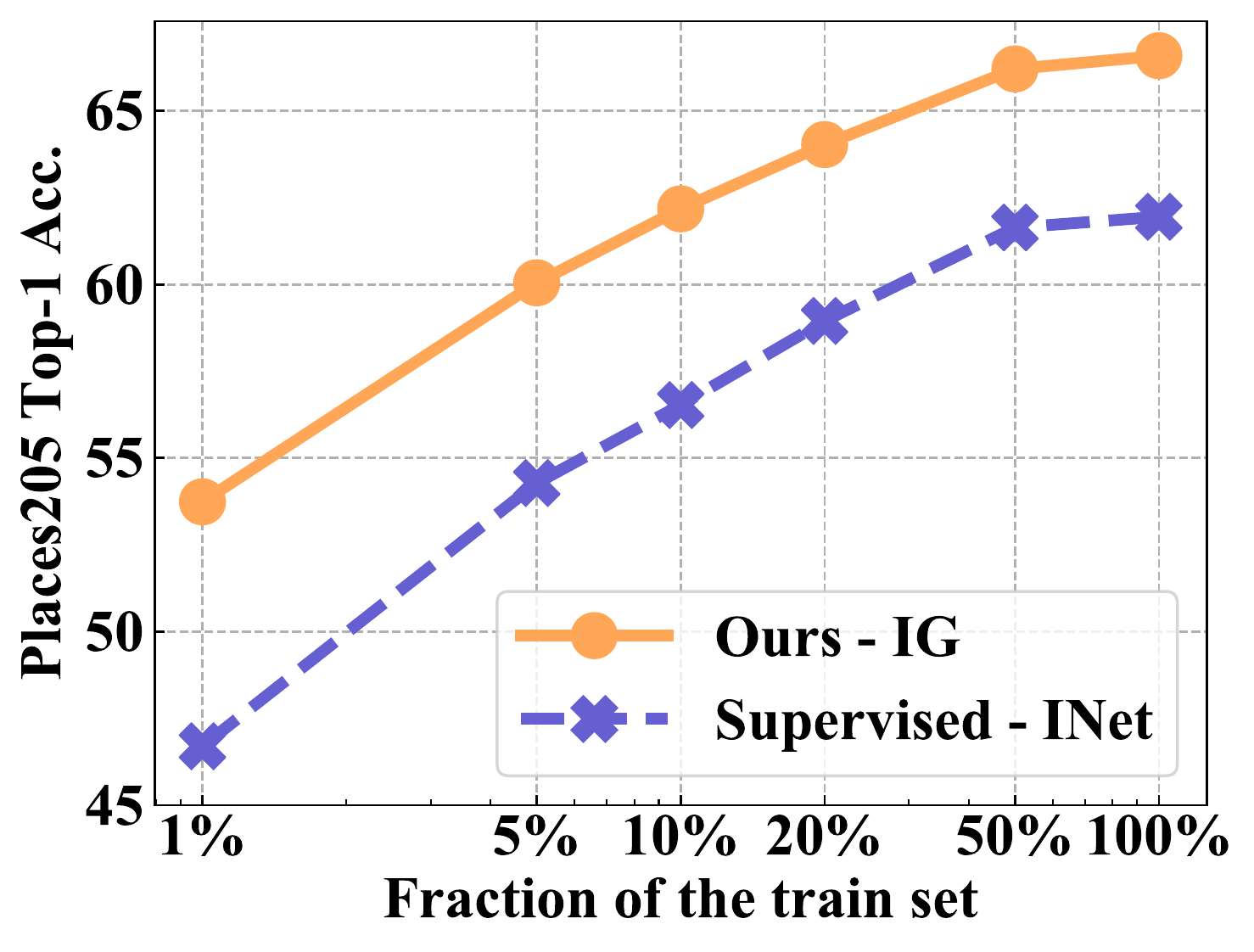}
  \caption{
    \textbf{Low-shot learning on Places.}
We compare the impact of dfferent pretraining when transferring to Places205 with different fraction of the train set available for finetuning. 
We report Top-1 accuracy and use a \texttt{RegNetY-128GF} architectures for our pretraining and the supervised pretraining on ImageNet.
  }
  \label{fig:semi_sup_places}
\end{figure}

\paragraph{Results on Places205.}
In Figure~\ref{fig:semi_sup_places}, we show the impact of pretraining on different fractions of the Places205 dataset~\cite{zhou2014learning}.
We compare to pretraining on ImageNet with supervision with the same \texttt{RegNetY-128GF} architecture.
A surprising result is that we observe a stable gain of $2.5\%$ in top-1 accuracy, regardless of the fraction of training data available to finetune on Places205.
The difference between self-supervised and supervised pretraining may be explained by the difference in the nature of training data: features learned from images in the wild may be more suitable to classify scene. 
Additionally, the non-uniform distribution of underlying concepts in the wild may also provide an advantage to our pretraining on a unbalanced dataset like Places205.

\paragraph{Results on ImageNet.}
In Table~\ref{tab:semi_sup_inet}, we show the performance of our self-supervised pretrained model on low-shot learning. 
For completeness, we report performance of existing semi-supervised and self-supervised methods.
We note that all of these methods use the \emph{entire} set of $1.2$M images from ImageNet for pretraining and only restrict the access to the labels, while we only see 1\% and 10\% of the images. 
This greatly favors these approaches since the network has seen more images from the same distribution during pretraining as the fraction used for transfer.
Nonetheless, our approach achieves a top-1 accuracy of 77.9\% with only 10\% of ImageNet, which is competitive with these methods (2\% gap).
On 1\% of the data, i.e, 10K images, the gap increases significantly but note that the other methods are using the full ImageNet from pretraining.

\begin{figure}[t]
  \centering
  \includegraphics[width=.92\linewidth]{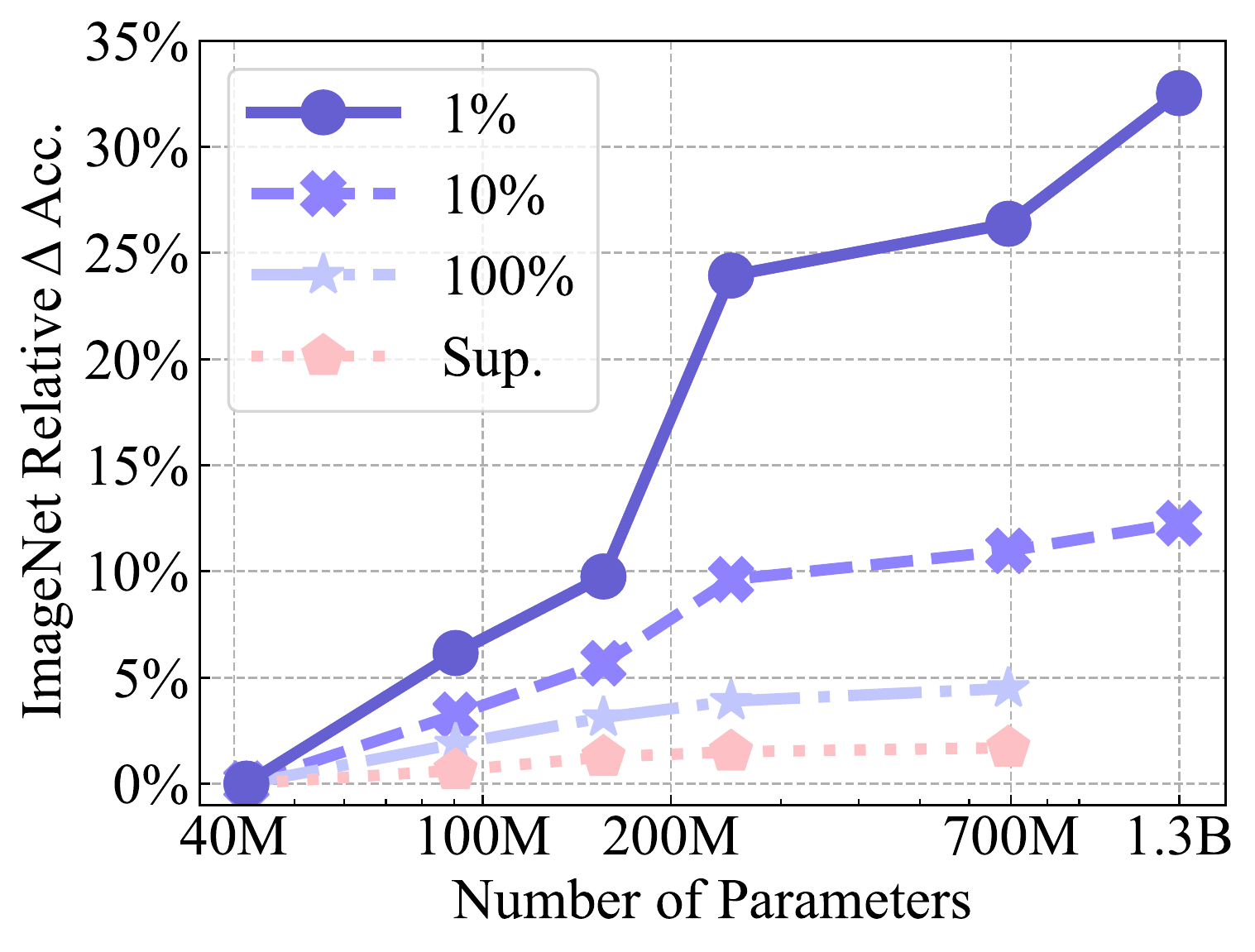}
  \caption{\textbf{Impact of capacity on low-shot learning.} 
We report the relative improvement in top-1 accuracy when finetuning pretrained RegNets with different capacities on a fraction of ImageNet.
Note that we only access to 1\% and 10\% of the images and their labels.
We also report the relative improvement for a pretrained model finetuned on the full ImageNet dataset (``100\%'').
For reference, we report the relative improvement of RegNets trained with supervision on the full ImageNet dataset (``Sup.'').
}
\label{fig:lowshot}
\end{figure}

\paragraph{Impact of the model capacity.}
In Figure~\ref{fig:lowshot}, we explore the impact of model capacity in the different low-shot settings - 1\%, 10\% and 100\% of ImageNet.
A first observation is that increasing model capacity gives a higher relative improvement as we decrease the access to both labels and images. 
This result extends the observation of Chen et al.~\cite{chen2020big} on the low-shot setting.
Interestingly, the relative gains are comparable in both settings ($+20\%$ in $1\%$ of the data), even though low-shot learning is strictly harder.

\begin{table}[t]
  \centering
  \begin{tabular}{@{}l  l cccc@{}}
    \toprule
   & Arch. & iNat. & OpIm. & Places & VOC  \\
\midrule
\multicolumn{6}{@{}l}{\small\textit{Existing pretrained features on ImageNet}}\\
SwAV~\cite{caron2020unsupervised}  & RN50 & 48.6 & 81.0 & 56.7 & 88.9\\
\midrule
\multicolumn{6}{@{}l}{\small\textit{Pretrained on ImageNet}}\\
Sup.  & RG128 & 47.2 & 81.1 & 56.0 & 89.2  \\
SwAV  & RG128 & 47.5 & 83.9 & 59.9 & 89.8 \\
\midrule
\multicolumn{6}{@{}l}{\small\textit{Pretrained on uncurated data}}\\
        SEER  & RG128   & 47.2 & 84.9 & 61.9 & 91.6  \\
    SEER  & RG256   & \textbf{50.8} & \textbf{85.8} & \textbf{62.7} & \textbf{92.6}  \\
\bottomrule
\end{tabular}
\vspace{.2em}
  \caption{
    \textbf{Linear Evaluation on downstream classification tasks.}
We compare the features from different pretrainings with a linear evaluation on top of frozen features.
We report accuracy on the following downstream tasks: iNaturalist (``iNat.'')~\cite{van2018inaturalist}, 
OpenImages (``OpIm.'')~\cite{kuznetsova2018open}, Places205~\cite{zhou2014learning} 
and Pascal VOC2007~\cite{everingham2010pascal}.
  }
  \label{tab:linear}
\end{table}

\subsection{Transfer to Other Benchmarks}

In these experiments, we further evaluate our pretrained features by transferring them to other downstream tasks.

\paragraph{Linear evaluation of image classification.}
In Table~\ref{tab:linear}, we compare the features from our pretrained \texttt{RegNetY-128GF} and \texttt{RegNetY-256GF} with features from the same architecture pretrained on ImageNet with and without supervision. 
To assess features quality, we freeze the model weights and learn a linear classifier on top of the features using the training set of each downstream task.
We consider the following benchmarks: iNaturalist~\cite{van2018inaturalist}, OpenImages~\cite{kuznetsova2018open}, Places205~\cite{zhou2014learning} and Pascal VOC~\cite{everingham2010pascal}.
We observe that self-supervised features transfer better than supervised features regardless of the pretraining data.

\paragraph{Detection and segmentation.}
In Table~\ref{tab:coco}, we evalaute pretrained features on detection and segmentation.
We train a Mask-RCNN model~\cite{he2017mask} on the COCO benchmark~\cite{lin2014microsoft} with pretrained \texttt{RegNetY-64GF} and \texttt{RegNetY-128GF} as backbones.
For both downstream tasks and architectures, our self-supervised pretraining \textit{\textbf{outperforms supervised pretraining by $1.5-2$ AP} points}.
However, the gap in performances between different architectures is small ($0.1-0.5$ AP) compared to what we observed on ImageNet.

\begin{table}[t]
\centering
\begin{tabular}{@{}l l l cc@{}}
\toprule
Method     & Data & Arch. &  AP$_{\text{box}}$ &  AP$_{\text{mask}}$ \\ 
\midrule
Supervised & INet & RG64  & 45.9 & 41.0 \\
Supervised & INet & RG128 & 46.6 & 41.6 \\
\midrule
SEER       & IG   & RG64  & 48.1 & 43.1 \\
SEER       & IG   & RG128 & \textbf{48.5} & \textbf{43.2} \\
\bottomrule
\end{tabular}
\vspace{.4em}
\caption{
\textbf{Detection and Segmentation on COCO.}
We compare the performance of Mask-RCNN models~\cite{he2017mask} initialized with different pretrained RegNet architectures as backbone on the detection and segmentation tasks of COCO~\cite{lin2014microsoft}. 
We consider two architectures, \texttt{RegNetY-64gf} and \texttt{RegNetY-128gf}, that we either pretrained with supervision on ImageNet or without supervision on 1B IG images.
}
\label{tab:coco}
\end{table}

\subsection{Comparing to Weakly-Supervised Pretraining}

Many online images have some metadata, e.g., hashtags or geo-localization, that can be leveraged during pretraining.
In particular, Mahajan et al.~\cite{mahajan2018exploring} show that pretraining by predicting a curated set of hashtags can greatly improve the quality of the resulting visual features.
Their approach requires to filter images and only works in the presence of textual metadata.
In Table~\ref{tab:weak}, we compare our self-supervised pretraining on \emph{random} images to theirs on the same architecture, a ResNeXt101-32x8d, with finetuning.
For completeness, we also report their best number with their largest architecture.
First, we observe that both pretrainings improve top-1 accuracy over a model trained from scratch, showing in general the benefits of pretraining.
Our approach is also in the same ballpark as theirs even though we do not rely on data curation nor supervision.
Note that, when the features are frozen, their approach maintains high performance on ImageNet, with $81.6\%$ top-1 accuracy while our model performance drops significantly -- around $65\%$ top-1.
This result is not surprising: they pretrain on data that follows the same concepts as ImageNet classes and thus the learned features are more aligned with the target distribution.
Since we pretrain our model on random images, we require a full-finetuning step of $35$ epochs to adapt to the target distribution.
This experiment shows that the benefits of pretraining with finetuning exist even if the features come from a different image distribution.  

\begin{table}[t]
\centering 
\begin{tabular}{@{}l l c c@{}}
\toprule
Pretraining & Arch. & \#param  & Top-1 \\
\midrule
\multicolumn{4}{@{}l}{\small\textit{Same architecture}}\\
Scratch & RX101-32x8d & \phantom{0}91M & 79.6 \\
Hashtag pred.~\cite{mahajan2018exploring} & RX101-32x8d & \phantom{0}91M & 82.6\\
SEER & RX101-32x8d & \phantom{0}91M   & 81.6 \\
\midrule
\multicolumn{4}{@{}l}{\small\textit{Different architectures}}\\
Hashtag pred.~\cite{mahajan2018exploring} & RX101-32x48d & 831M & 85.4\\
SEER & RG128 & 693M   & 83.8 \\
SEER & RG256 & 1.3B   & 84.2 \\
\bottomrule
\end{tabular}
\vspace{.3em}
\caption{\textbf{Comparision with weakly-supervised pretraining on curated data.}
We compare pretraining a ResNeXt101-32dx8d with self-supervision on \emph{random} images with pretraining on filtered images \emph{labeled} with hashtags that are similar to ImageNet classes~\cite{mahajan2018exploring}.
We report top-1 accuracy on ImageNet with finetuning.
For completeness, we also report the best performance reported with larger architectures.
}
\label{tab:weak}
\end{table}

\begin{figure}[h]
  \centering
  \includegraphics[width=.92\linewidth]{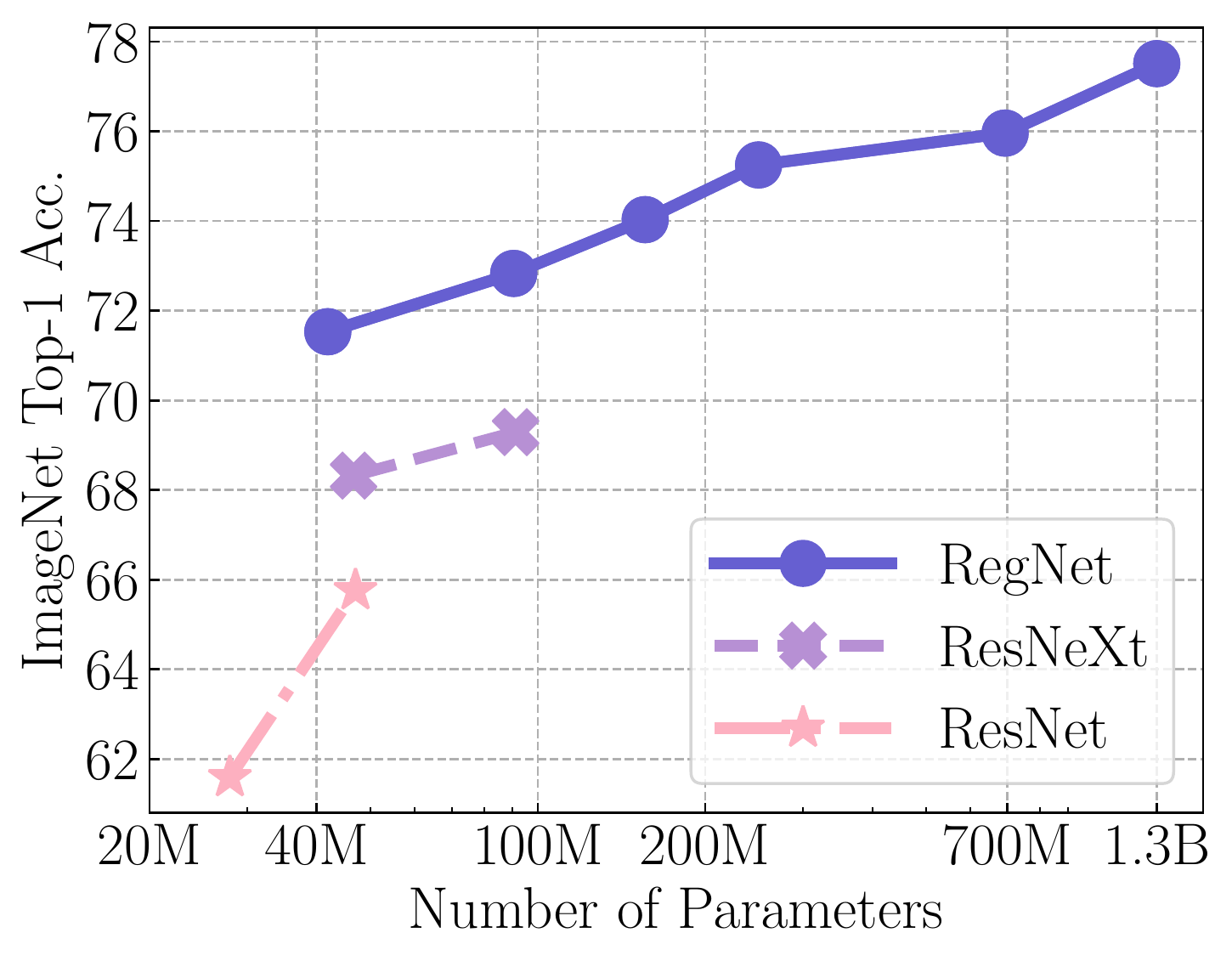}
  \caption{\textbf{Comparison across architectures.} 
We pretrain different ResNets, ResNeXts and RegNetYs for $1$ epoch on $1$B IG images with SwAV.
We report top-1 accuracy on ImageNet of a linear classifier trained on frozen features.}
\label{fig:model}
\end{figure}

\begin{figure*}[t]
\begin{tabular}{cc}
  \includegraphics[width=.45\linewidth]{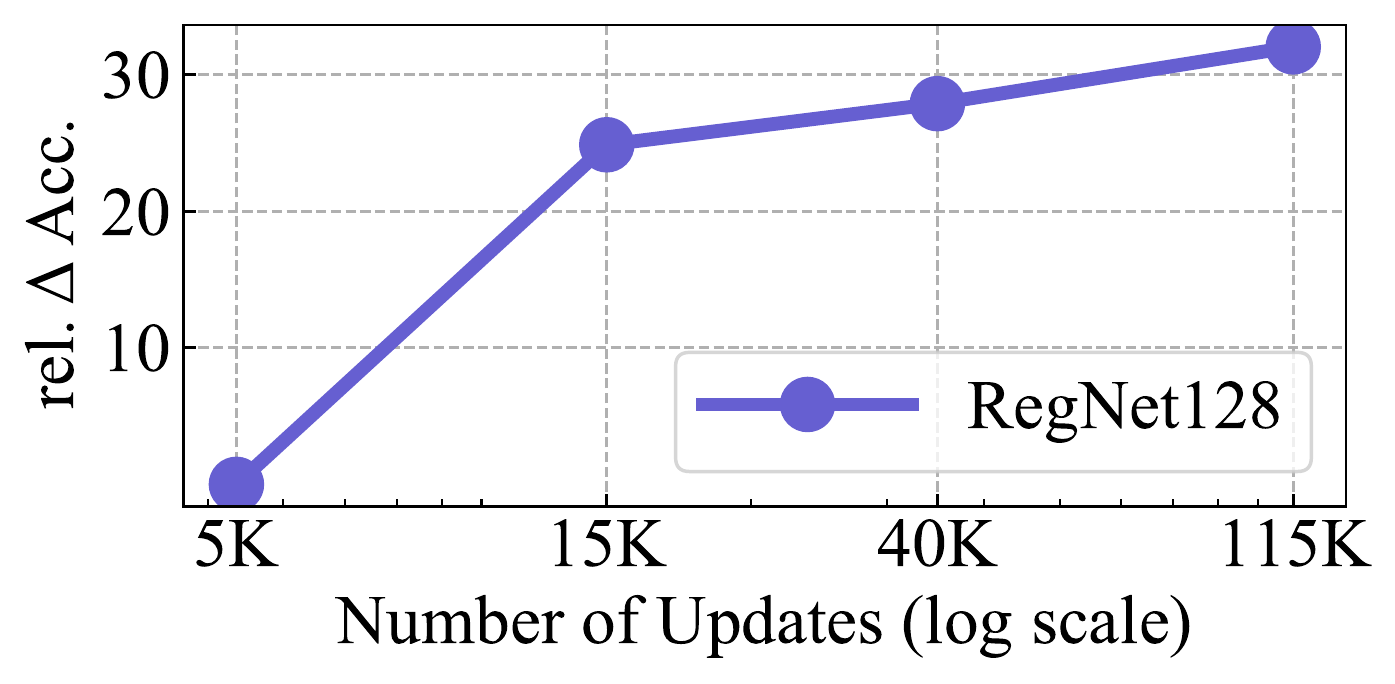}&
  \includegraphics[width=.45\linewidth]{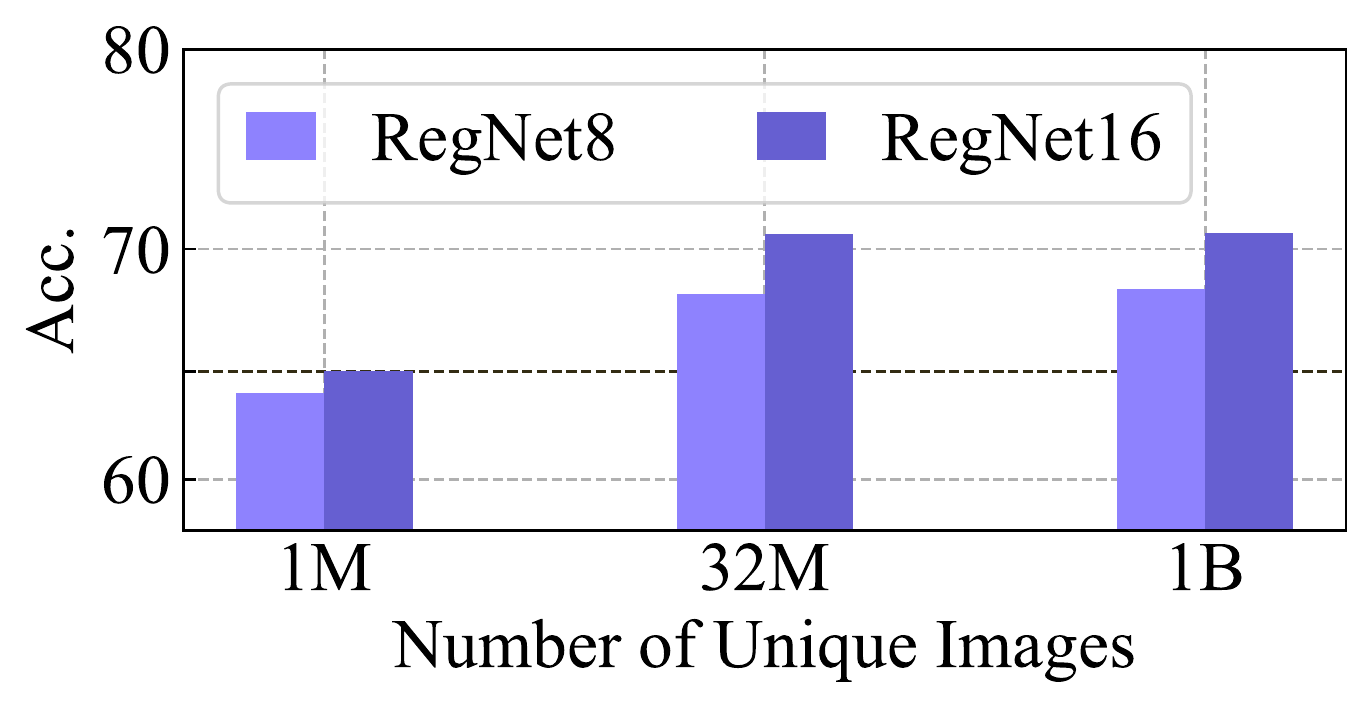} 
\end{tabular}
  \caption{
\textbf{(left) Impact of number of updates.}
We compare the quality of a \texttt{RegNetY-128GF} after different number of updates of an online pretraining on 1B images.
For both studies, we report the relative improvement in top-1 accuracy for a linear evaluation of frozen features on ImageNet.
\textbf{(right) Impact of number of unique images.}
We compare the impact of the size of the training set for a \texttt{RegNetY-8GF} and a \texttt{RegNetY-16GF} pretrained for the same number of updates.
The number of updates corresponds to $1$ epoch for $1$B images, $32$ epochs for $32$M images and $1$K for $1$M images.
}
\label{fig:data}
\end{figure*}

\section{Ablation Studies}

These ablation studies focus on the model architecture, how its performance scales with capacity, and the specificities of our pretraining data and our self-supervised method.

\subsection{Impact of the Model Architecture}

\paragraph{Experimental setting.} 
We consider several RegNetY architectures with growing capacity, namely the \texttt{RegNetY-\{8,16,32,64,128\}GF}.
We also consider ResNet-\{50, 101\}~\cite{he2016deep} and the ResNeXt architectures with a growing number of parameters, namely RX101-32x\{4, 8\}d~\cite{xie2017aggregated}.
We refer to the appendix for details about the different architectures. 
Every model is pretrained for $1$ epoch on $1$B random, public and non-EU Instagram (IG) images with SwAV using $3$K prototypes. We use same hyperparameters for pretraining all the ablation models.

\paragraph{Impact of the architecture.}
In Figure~\ref{fig:model}, we measure how changing the architecture affects the quality of pretrained features with a linear evaluation on ImageNet.
This evaluation does not favor models that perform well when trained from scratch on ImageNet, and hence, we directly probe the pretrained features. 
For ResNets and ResNeXts, we observe that the features from the penultimate layer work better in this setting and we report those for fair comparison with RegNets.
Overall, RegNets surpass the other architectures, justifying our choice of architecture for our main model.
Finally, we observe, regardless of the architecture, increasing model capacity significantly improves the quality of the features with a logarithmic gain in performance.

\subsection{Scaling the Training Data}

Pretraining on a larger dataset can improve the quality of the learned visual features for two reasons: more parameter updates and more unique images.
In this section, we disentangle these two effects.

\paragraph{Increasing the number of updates.} 
On the left panel of Figure~\ref{fig:data}, we show the performance as we train a \texttt{RegNetY-128GF} model online on 1B images.
We observe that the performance steadily increases with the number of updates as expected and the performance does not saturate even after a number of updates corresponding to $1$B images.

\paragraph{Increasing the number of unique images.}
On the right panel of Figure~\ref{fig:data}, we report the performance of two models, \texttt{RegNetY-8GF} and \texttt{RegNetY-16GF} when trained for the same number of updates but with a different number of unique images.
We train the models for a number of updates that corresponds to $1$ epoch over $1$B unique images, or $32$ epochs for $32$M unique images, with a single half-cosine wave learning rate.
An interesting observation is that, with this learning rate schedule, the minimum number of unique images required to obtain good performance is greater than the size of ImageNet by \emph{only} an order of magnitude. 

Overall, these experiments show that the number of updates matters more than seeing the same images multiple times.
There is thus no need to fix the pretraining dataset, and instead validating their continual online pretraining.

\subsection{Scaling the self-supervised model head}

In this section, we study the impact of growing the size of the self-supervised model head during pretraining.

In Table~\ref{tab:swav}, we compare \texttt{RegNetY-8GF} architectures trained with different capacity self-supervised heads.
We report top-1 accuracy on ImageNet obtained with a linear classifier trained on frozen features.
The models are trained on $1$B images with a cosine wave learning rate schedule.
In particular, we show the impact of a larger MLP and more prototype vectors.
We adjust the head from 2-layer MLP of dimensions ($2016\!\times\!2016$ and $2016\!\times\!256$) to 3-layer MLP of dimensions ($2016\!\times\!4096$, $4096\!\times\!4096$, $4096\!\times\!256$), and increase the number of prototypes from $3$K to $16$K.
We observe that simply increasing the number of the parameters in the head, and the number of clusters significantly improves the performance of the resulting model ($+3$\%) with the same model, hence same feature size.
The reason is that $1$B random images contain much more concepts that the original SwAV classifier can memorize in its small head, hence the information about the clusters leaks to the features, degrading their performance.
Increasing the head reduces this effect at a minimal compute and storage cost.

\begin{table}[t]
\centering
\begin{tabular}{@{}l cc cc cc@{}}
\toprule
&&& \multicolumn{2}{c}{Top-1}\\
\cmidrule{4-5}
Head & Dimension of MLP & \#proto& res4 & res5  \\
\midrule
Small & $[2016,2016,256]$ & $3$K & 68.8 & 68.1 \\
Large & $[2016,4096,4096,256]$ & $16$K & 67.1 & \textbf{71.5} \\
\bottomrule
\end{tabular}
\vspace{.3em}
\caption{\textbf{Adapting SwaV to IG data.}
We show the impact of scaling the head of our self-supervised loss, by increasing the size of the MLP and the number of prototypes.
We report top-1 accuracy on ImageNet with a linear evaluation on top of the features from the last (``res5'') and penultimate blocks (``res4'') of a \texttt{RegNetY-8GF} architectures pretrained on $1$B random public and non-EU IG images.
}
\label{tab:swav}
\end{table}

\section{Conclusion}
We show that pretraining features on random images with no annotation achieves competitive performance on any downstream task.
This result confirm that the recent progress of self-supervised learning is not specific to curated training set, like ImageNet, and could benefit a large range of applications associated with uncurated data.
Our work benefits from the scalability of modern self-supervised learning methods in terms of data, and modern efficient high-capacity architectures. 
In particular, the scalability of RegNets have played a key role in pushing the limits of self-supervised pretraining, and in the future, we plan to search for larger RegNet architectures suited for this task. 

{\small
\bibliographystyle{ieee_fullname}
\bibliography{egbib}

\begin{thebibliography}{10}\itemsep=-1pt

\bibitem{asano2019self}
Yuki~Markus Asano, Christian Rupprecht, and Andrea Vedaldi.
\newblock Self-labelling via simultaneous clustering and representation
  learning.
\newblock In {\em Proceedings of the International Conference on Learning
  Representations (ICLR)}, 2020.

\bibitem{baevski2020wav2vec}
Alexei Baevski, Henry Zhou, Abdelrahman Mohamed, and Michael Auli.
\newblock wav2vec 2.0: A framework for self-supervised learning of speech
  representations.
\newblock In {\em Proceedings of Advances in Neural Information Processing
  Systems (NeurIPS)}, 2020.

\bibitem{bojanowski2017unsupervised}
Piotr Bojanowski and Armand Joulin.
\newblock Unsupervised learning by predicting noise.
\newblock In {\em Proceedings of the International Conference on Machine
  Learning (ICML)}, 2017.

\bibitem{brown2020language}
Tom~B Brown, Benjamin Mann, Nick Ryder, Melanie Subbiah, Jared Kaplan, Prafulla
  Dhariwal, Arvind Neelakantan, Pranav Shyam, Girish Sastry, Amanda Askell,
  et~al.
\newblock Language models are few-shot learners.
\newblock {\em arXiv preprint arXiv:2005.14165}, 2020.

\bibitem{caron2018deep}
Mathilde Caron, Piotr Bojanowski, Armand Joulin, and Matthijs Douze.
\newblock Deep clustering for unsupervised learning of visual features.
\newblock In {\em Proceedings of the European Conference on Computer Vision
  (ECCV)}, 2018.

\bibitem{caron2019unsupervised}
Mathilde Caron, Piotr Bojanowski, Julien Mairal, and Armand Joulin.
\newblock Unsupervised pre-training of image features on non-curated data.
\newblock In {\em Proceedings of the International Conference on Computer
  Vision (ICCV)}, 2019.

\bibitem{caron2020unsupervised}
Mathilde Caron, Ishan Misra, Julien Mairal, Priya Goyal, Piotr Bojanowski, and
  Armand Joulin.
\newblock Unsupervised learning of visual features by contrasting cluster
  assignments.
\newblock In {\em Proceedings of Advances in Neural Information Processing
  Systems (NeurIPS)}, 2020.

\bibitem{chen2020simple}
Ting Chen, Simon Kornblith, Mohammad Norouzi, and Geoffrey Hinton.
\newblock A simple framework for contrastive learning of visual
  representations.
\newblock {\em arXiv preprint arXiv:2002.05709}, 2020.

\bibitem{chen2020big}
Ting Chen, Simon Kornblith, Kevin Swersky, Mohammad Norouzi, and Geoffrey
  Hinton.
\newblock Big self-supervised models are strong semi-supervised learners.
\newblock In {\em Proceedings of Advances in Neural Information Processing
  Systems (NeurIPS)}, 2020.

\bibitem{chen2016grad}
Tianqi Chen, Bing Xu, Chiyuan Zhang, and Carlos Guestrin.
\newblock Training deep nets with sublinear memory cost.
\newblock {\em arXiv preprint arXiv:1604.06174}, 2016.

\bibitem{coates2011analysis}
Adam Coates, Andrew Ng, and Honglak Lee.
\newblock An analysis of single-layer networks in unsupervised feature
  learning.
\newblock In {\em Proceedings of the International Conference on Artificial
  Intelligence and Statistics (AISTATS)}, 2011.

\bibitem{devlin2018bert}
Jacob Devlin, Ming-Wei Chang, Kenton Lee, and Kristina Toutanova.
\newblock Bert: Pre-training of deep bidirectional transformers for language
  understanding.
\newblock {\em arXiv preprint arXiv:1810.04805}, 2018.

\bibitem{doersch2015unsupervised}
Carl Doersch, Abhinav Gupta, and Alexei~A Efros.
\newblock Unsupervised visual representation learning by context prediction.
\newblock In {\em Proceedings of the International Conference on Computer
  Vision (ICCV)}, 2015.

\bibitem{dosovitskiy2020image}
Alexey Dosovitskiy, Lucas Beyer, Alexander Kolesnikov, Dirk Weissenborn,
  Xiaohua Zhai, Thomas Unterthiner, Mostafa Dehghani, Matthias Minderer, Georg
  Heigold, Sylvain Gelly, et~al.
\newblock An image is worth 16x16 words: Transformers for image recognition at
  scale.
\newblock {\em arXiv preprint arXiv:2010.11929}, 2020.

\bibitem{dosovitskiy2016discriminative}
Alexey Dosovitskiy, Philipp Fischer, Jost~Tobias Springenberg, Martin
  Riedmiller, and Thomas Brox.
\newblock Discriminative unsupervised feature learning with exemplar
  convolutional neural networks.
\newblock {\em Transactions on Pattern Analysis and Machine Intelligence
  (TPAMI)}, 2016.

\bibitem{everingham2010pascal}
Mark Everingham, Luc Van~Gool, Christopher~KI Williams, John Winn, and Andrew
  Zisserman.
\newblock The pascal visual object classes (voc) challenge.
\newblock {\em International Journal of Computer Vision (IJCV)}, 2010.

\bibitem{french2020milking}
Geoff French, Avital Oliver, and Tim Salimans.
\newblock Milking cowmask for semi-supervised image classification.
\newblock {\em arXiv preprint arXiv:2003.12022}, 2020.

\bibitem{goyal2017accurate}
Priya Goyal, Piotr Doll{\'a}r, Ross Girshick, Pieter Noordhuis, Lukasz
  Wesolowski, Aapo Kyrola, Andrew Tulloch, Yangqing Jia, and Kaiming He.
\newblock Accurate, large minibatch sgd: Training imagenet in 1 hour.
\newblock {\em arXiv preprint arXiv:1706.02677}, 2017.

\bibitem{goyal2019scaling}
Priya Goyal, Dhruv Mahajan, Abhinav Gupta, and Ishan Misra.
\newblock Scaling and benchmarking self-supervised visual representation
  learning.
\newblock In {\em Proceedings of the International Conference on Computer
  Vision (ICCV)}, 2019.

\bibitem{grill2020bootstrap}
Jean-Bastien Grill, Florian Strub, Florent Altch{\'e}, Corentin Tallec,
  Pierre~H Richemond, Elena Buchatskaya, Carl Doersch, Bernardo~Avila Pires,
  Zhaohan~Daniel Guo, Mohammad~Gheshlaghi Azar, et~al.
\newblock Bootstrap your own latent: A new approach to self-supervised
  learning.
\newblock In {\em Proceedings of Advances in Neural Information Processing
  Systems (NeurIPS)}, 2020.

\bibitem{hadsell2006dimensionality}
Raia Hadsell, Sumit Chopra, and Yann LeCun.
\newblock Dimensionality reduction by learning an invariant mapping.
\newblock In {\em Proceedings of the Conference on Computer Vision and Pattern
  Recognition (CVPR)}, 2006.

\bibitem{he2020momentum}
Kaiming He, Haoqi Fan, Yuxin Wu, Saining Xie, and Ross Girshick.
\newblock Momentum contrast for unsupervised visual representation learning.
\newblock In {\em Proceedings of the Conference on Computer Vision and Pattern
  Recognition (CVPR)}, 2020.

\bibitem{he2017mask}
Kaiming He, Georgia Gkioxari, Piotr Doll{\'a}r, and Ross Girshick.
\newblock Mask r-cnn.
\newblock In {\em Proceedings of the International Conference on Computer
  Vision (ICCV)}, 2017.

\bibitem{he2016deep}
Kaiming He, Xiangyu Zhang, Shaoqing Ren, and Jian Sun.
\newblock Deep residual learning for image recognition.
\newblock In {\em Proceedings of the Conference on Computer Vision and Pattern
  Recognition (CVPR)}, 2016.

\bibitem{henaff2019data}
Olivier~J H{\'e}naff, Aravind Srinivas, Jeffrey De~Fauw, Ali Razavi, Carl
  Doersch, SM Eslami, and Aaron van~den Oord.
\newblock Data-efficient image recognition with contrastive predictive coding.
\newblock {\em arXiv preprint arXiv:1905.09272}, 2019.

\bibitem{hu2018squeeze}
Jie Hu, Li Shen, and Gang Sun.
\newblock Squeeze-and-excitation networks.
\newblock In {\em Proceedings of the Conference on Computer Vision and Pattern
  Recognition (CVPR)}, 2018.

\bibitem{joulin2016learning}
Armand Joulin, Laurens Van Der~Maaten, Allan Jabri, and Nicolas Vasilache.
\newblock Learning visual features from large weakly supervised data.
\newblock In {\em Proceedings of the European Conference on Computer Vision
  (ECCV)}, 2016.

\bibitem{kahn2020libri}
Jacob Kahn, Morgane Rivi{\`e}re, Weiyi Zheng, Evgeny Kharitonov, Qiantong Xu,
  Pierre-Emmanuel Mazar{\'e}, Julien Karadayi, Vitaliy Liptchinsky, Ronan
  Collobert, Christian Fuegen, et~al.
\newblock Libri-light: A benchmark for asr with limited or no supervision.
\newblock In {\em Proceedings of the International Conference on Acoustics,
  Speech and Signal Processing (ICASSP)}, 2020.

\bibitem{kolesnikov2019big}
Alexander Kolesnikov, Lucas Beyer, Xiaohua Zhai, Joan Puigcerver, Jessica Yung,
  Sylvain Gelly, and Neil Houlsby.
\newblock Big transfer (bit): General visual representation learning.
\newblock In {\em Proceedings of the European Conference on Computer Vision
  (ECCV)}, 2020.

\bibitem{kolesnikov2019revisiting}
Alexander Kolesnikov, Xiaohua Zhai, and Lucas Beyer.
\newblock Revisiting self-supervised visual representation learning.
\newblock In {\em Proceedings of the Conference on Computer Vision and Pattern
  Recognition (CVPR)}, 2019.

\bibitem{kuznetsova2018open}
Alina Kuznetsova, Mohamad Hassan~Mohamad Rom, Neil Alldrin, Jasper Uijlings,
  Ivan Krasin, Jordi Pont-Tuset, Shahab Kamali, Stefan Popov, Matteo Malloci,
  Alexander Kolesnikov, Tom Duerig, and Vittorio Ferrari.
\newblock The open images dataset v4: Unified image classification, object
  detection, and visual relationship detection at scale.
\newblock {\em International Journal of Computer Vision (IJCV)}, 2020.

\bibitem{li2020prototypical}
Junnan Li, Pan Zhou, Caiming Xiong, Richard Socher, and Steven~CH Hoi.
\newblock Prototypical contrastive learning of unsupervised representations.
\newblock {\em arXiv preprint arXiv:2005.04966}, 2020.

\bibitem{lin2014microsoft}
Tsung-Yi Lin, Michael Maire, Serge Belongie, James Hays, Pietro Perona, Deva
  Ramanan, Piotr Doll{\'a}r, and C~Lawrence Zitnick.
\newblock Microsoft coco: Common objects in context.
\newblock In {\em Proceedings of the European Conference on Computer Vision
  (ECCV)}, 2014.

\bibitem{loshchilov2016sgdr}
Ilya Loshchilov and Frank Hutter.
\newblock Sgdr: Stochastic gradient descent with warm restarts.
\newblock {\em arXiv preprint arXiv:1608.03983}, 2016.

\bibitem{mahajan2018exploring}
Dhruv Mahajan, Ross Girshick, Vignesh Ramanathan, Kaiming He, Manohar Paluri,
  Yixuan Li, Ashwin Bharambe, and Laurens van~der Maaten.
\newblock Exploring the limits of weakly supervised pretraining.
\newblock In {\em Proceedings of the European Conference on Computer Vision
  (ECCV)}, 2018.

\bibitem{miech2020end}
Antoine Miech, Jean-Baptiste Alayrac, Lucas Smaira, Ivan Laptev, Josef Sivic,
  and Andrew Zisserman.
\newblock End-to-end learning of visual representations from uncurated
  instructional videos.
\newblock In {\em Proceedings of the Conference on Computer Vision and Pattern
  Recognition (CVPR)}, 2020.

\bibitem{misra2020self}
Ishan Misra and Laurens van~der Maaten.
\newblock Self-supervised learning of pretext-invariant representations.
\newblock In {\em Proceedings of the Conference on Computer Vision and Pattern
  Recognition (CVPR)}, 2020.

\bibitem{oord2018representation}
Aaron van~den Oord, Yazhe Li, and Oriol Vinyals.
\newblock Representation learning with contrastive predictive coding.
\newblock {\em arXiv preprint arXiv:1807.03748}, 2018.

\bibitem{radford2019language}
Alec Radford, Jeffrey Wu, Rewon Child, David Luan, Dario Amodei, and Ilya
  Sutskever.
\newblock Language models are unsupervised multitask learners.

\bibitem{radosavovic2020designing}
Ilija Radosavovic, Raj~Prateek Kosaraju, Ross Girshick, Kaiming He, and Piotr
  Doll{\'a}r.
\newblock Designing network design spaces.
\newblock In {\em Proceedings of the Conference on Computer Vision and Pattern
  Recognition (CVPR)}, 2020.

\bibitem{raffel2019exploring}
Colin Raffel, Noam Shazeer, Adam Roberts, Katherine Lee, Sharan Narang, Michael
  Matena, Yanqi Zhou, Wei Li, and Peter~J Liu.
\newblock Exploring the limits of transfer learning with a unified text-to-text
  transformer.
\newblock {\em arXiv preprint arXiv:1910.10683}, 2019.

\bibitem{ranzato2007unsupervised}
Marc’Aurelio Ranzato, Fu-Jie Huang, Y-Lan Boureau, and Yann LeCun.
\newblock Unsupervised learning of invariant feature hierarchies with
  applications to object recognition.
\newblock In {\em Proceedings of the Conference on Computer Vision and Pattern
  Recognition (CVPR)}, 2007.

\bibitem{riviere2020unsupervised}
Morgane Rivi{\`e}re, Armand Joulin, Pierre-Emmanuel Mazar{\'e}, and Emmanuel
  Dupoux.
\newblock Unsupervised pretraining transfers well across languages.
\newblock In {\em Proceedings of the International Conference on Acoustics,
  Speech and Signal Processing (ICASSP)}, 2020.

\bibitem{russakovsky2015imagenet}
Olga Russakovsky, Jia Deng, Hao Su, Jonathan Krause, Sanjeev Satheesh, Sean Ma,
  Zhiheng Huang, Andrej Karpathy, Aditya Khosla, Michael Bernstein, Alexander~C
  Berg, and Li Fei-Fei.
\newblock Imagenet large scale visual recognition challenge.
\newblock {\em International Journal of Computer Vision (IJCV)}, 2015.

\bibitem{schneider2019wav2vec}
Steffen Schneider, Alexei Baevski, Ronan Collobert, and Michael Auli.
\newblock wav2vec: Unsupervised pre-training for speech recognition.
\newblock {\em arXiv preprint arXiv:1904.05862}, 2019.

\bibitem{sohn2020fixmatch}
Kihyuk Sohn, David Berthelot, Chun-Liang Li, Zizhao Zhang, Nicholas Carlini,
  Ekin~D Cubuk, Alex Kurakin, Han Zhang, and Colin Raffel.
\newblock Fixmatch: Simplifying semi-supervised learning with consistency and
  confidence.
\newblock In {\em Proceedings of Advances in Neural Information Processing
  Systems (NeurIPS)}, 2020.

\bibitem{sun2017revisiting}
Chen Sun, Abhinav Shrivastava, Saurabh Singh, and Abhinav Gupta.
\newblock Revisiting unreasonable effectiveness of data in deep learning era.
\newblock In {\em Proceedings of the International Conference on Computer
  Vision (ICCV)}, 2017.

\bibitem{tan2019efficientnet}
Mingxing Tan and Quoc~V Le.
\newblock Efficientnet: Rethinking model scaling for convolutional neural
  networks.
\newblock {\em arXiv preprint arXiv:1905.11946}, 2019.

\bibitem{touvron2020fixing}
Hugo Touvron, Andrea Vedaldi, Matthijs Douze, and Herv{\'e} J{\'e}gou.
\newblock Fixing the train-test resolution discrepancy: Fixefficientnet.
\newblock {\em arXiv preprint arXiv:2003.08237}, 2020.

\bibitem{van2018inaturalist}
Grant Van~Horn, Oisin Mac~Aodha, Yang Song, Yin Cui, Chen Sun, Alex Shepard,
  Hartwig Adam, Pietro Perona, and Serge Belongie.
\newblock The inaturalist species classification and detection dataset.
\newblock In {\em Proceedings of the Conference on Computer Vision and Pattern
  Recognition (CVPR)}, 2018.

\bibitem{vincent2008extracting}
P. Vincent, H. Larochelle, Y. Bengio, and P.-A. Manzagol.
\newblock Extracting and composing robust features with denoising autoencoders.
\newblock In {\em Proceedings of the International Conference on Machine
  Learning (ICML)}, 2008.

\bibitem{wu2018unsupervised}
Zhirong Wu, Yuanjun Xiong, Stella~X Yu, and Dahua Lin.
\newblock Unsupervised feature learning via non-parametric instance
  discrimination.
\newblock In {\em Proceedings of the Conference on Computer Vision and Pattern
  Recognition (CVPR)}, 2018.

\bibitem{xie2017aggregated}
Saining Xie, Ross Girshick, Piotr Doll{\'a}r, Zhuowen Tu, and Kaiming He.
\newblock Aggregated residual transformations for deep neural networks.
\newblock In {\em Proceedings of the Conference on Computer Vision and Pattern
  Recognition (CVPR)}, 2017.

\bibitem{yan2020clusterfit}
Xueting Yan, Ishan Misra, Abhinav Gupta, Deepti Ghadiyaram, and Dhruv Mahajan.
\newblock Clusterfit: Improving generalization of visual representations.
\newblock In {\em Proceedings of the Conference on Computer Vision and Pattern
  Recognition (CVPR)}, 2020.

\bibitem{you2017large}
Yang You, Igor Gitman, and Boris Ginsburg.
\newblock Large batch training of convolutional networks.
\newblock {\em arXiv preprint arXiv:1708.03888}, 2017.

\bibitem{zhou2014learning}
Bolei Zhou, Agata Lapedriza, Jianxiong Xiao, Antonio Torralba, and Aude Oliva.
\newblock Learning deep features for scene recognition using places database.
\newblock In {\em Proceedings of Advances in Neural Information Processing
  Systems (NeurIPS)}, 2014.

\end{thebibliography}
}

\newpage
\section*{Supplementary Material}
\section{Model architectures.}
We describe below the model architecture settings used for ablation studies. In order to compare the architectures fairly, we follow the same hyperparameters for pre-training. We describe next the setup used for pretraining of ResNet-\{50,101\}, ResNeXt101-32x\{4,8\}d and \texttt{RegNetY-\{8,16,32,64,128\}GF}.

\subsection{Pretraining of ResNet and ResNeXt.}

We pretrain standard ResNet-\{50,101\} from He et al.~\cite{he2016deep} and standard RX101-32x\{4,8\}d from Xie et al.~\cite{xie2017aggregated} with SwAV, using $8$ crops per image of resolutions $2\times224 + 6\times96$. 
We follow the same data augmentation as in Caron et al.~\cite{caron2020unsupervised}.
During pretraining, we use a 2-layer multi-layer perceptron (MLP) projection head of dimensions $2048\times2048$ and $2048\times256$. We do not use BatchNorm layers in the head.
We use $3$K prototypes, temperature $\tau$ set to $0.1$, the Sinkhorn regularization parameter $\epsilon$ to $0.05$ and perform $5$ iterations of Sinkhorn algorithm.
We synchronize BatchNorm stats across gpus and create process groups of size 32 for synchronization.
We use a weight decay of $10^{-6}$, LARS optimizer~\cite{you2017large} and \texttt{O1} mixed-precision optimization from Apex library\footnote{\url{https://github.com/NVIDIA/apex}}. 
We train our model with stochastic gradient descent using a large batch size of $8192$ different images distributed over $256$ NVIDIA V100 32GB GPUs, resulting in $32$ different images per GPU. 
The learning rate is linearly ramped up from $0.3$ to $9.6$ for the first $6$K training updates. 
After warmup, we follow a half cosine wave learning rate schedule and decay the learning rate from $9.6$ to final value $0.0096$.
Overall, we train on $1$B images for a total of $122$K iterations.

\subsection{Pretraining of RegNet architectures.}
We train 5 different RegNet architectures namely the \texttt{RegNetY-\{8,16,32,64,128\}GF} of different capacity. RegNet architectures are generated by following the scaling rules described in Radosavovic et al.~\cite{radosavovic2020designing}. We first share the parametrization used for each of the RegNet architecture below. We then describe how we train these architectures with SwAV for our ablation study in Section 5.

\paragraph{RegNetY-8GF:} The model has \text{depth} = $17$ and RegNet parameters:
\begin{align*}
\small 
	w_0 = 192, w_a = 76.82, w_m = 2.19, \text{group width} = 56
\end{align*}

\paragraph{RegNetY-16GF.} The model has \text{depth} = $18$ and RegNet parameters:
\begin{align*}
\small
    w_0 = 200, w_a = 160.23, w_m = 2.48, \text{group width} = 112
\end{align*}

\paragraph{RegNetY-32GF.} The model has \text{depth} = $20$ and RegNet parameters:
\begin{align*}
\small
    w_0 = 232, w_a = 115.89, w_m = 2.53, \text{group width} = 232
\end{align*}

\paragraph{RegNetY-64GF.} The model has \text{depth} = $20$ and RegNet parameters:
\begin{align*}
\small
    w_0 = 352, w_a = 147.48, w_m = 2.4, \text{group width} = 328
\end{align*}

\paragraph{RegNetY-128GF.} The model has \text{depth} = $27$ and RegNet parameters:
\begin{align*}
\small
    w_0 = 456, w_a = 160.83, w_m = 2.52, \text{group width} = 264
\end{align*}

\paragraph{RegNetY-256GF.} The model has \text{depth} = $27$ and RegNet parameters:
\begin{align*}
\small
    w_0 = 640, w_a = 230.83, w_m = 2.53, \text{group width} = 373
\end{align*}

For pretraining the above RegNetY architectures, we follow the same pretraining hyperparams as ResNet and ResNeXt training with two differences. We use crops per image of resolutions $2\times224 + 4\times96$. However, we confirm that the crop resolutions didn't impact the model performance on ImageNet linear classification task on which we show our ablations in Section 5. Infact, using the bigger resolution crops leads to more GPU memory requirement with no impact on model performance on transfer task. The only other difference is the dimensions of 3-layer MLP in the head. Each RegNetY architecture has difference output channel and hence we adapt 3-layer MLP according to the architecture. More concretely, the head dimensions are: RegNetY-8GF has [$2016\times4096$, $4096\times4096$ and $4096\times256$], RegNetY-16GF has [$3024\times4096$, $4096\times4096$ and $4096\times256$], RegNetY-32GF has [$3712\times4096$, $4096\times4096$ and $4096\times256$], RegNetY-64GF has [$4920\times8192$, $8192\times8192$ and $8192\times256$], RegNetY-128GF has [$7392\times8192$, $8192\times8192$ and $8192\times256$]

\end{document}